\documentclass[10pt]{article} 
\usepackage[preprint]{tmlr}   
\usepackage{amsmath,amssymb,amsfonts}
\usepackage{booktabs}
\usepackage{array}
\usepackage{graphicx}
\usepackage{subcaption}
\usepackage[hidelinks]{hyperref}
\usepackage{url}
\usepackage{pifont}

\author{\name Yuxu Ge \email research@yuxu.ge \\
\addr University of York \\
\addr ORCID: \href{https://orcid.org/0009-0008-2990-4886}{0009-0008-2990-4886}}
\begin{document}
\title{Coverage, Not Credit: Failure-Credit Routing of Zeroth-Order
Perturbation Budgets Does Not Improve On-Pool Sample Efficiency for LLM
Agents}

\def\month{08}
\def\year{2026}
\def\openreview{\url{https://openreview.net/forum?id=TODO}}

\maketitle
\begin{abstract}
Trajectory-level credit assignment can localize which module of a
tool-using LLM agent causes failures using only verifiable signals. A
natural hypothesis is that such \emph{failure credit} should decide
where zeroth-order / evolution-strategies (ZO/ES) parameter search
spends its perturbation budget. We test this on a synthetic environment
and on frozen Qwen2.5-1.5B/3B and SmolLM2-1.7B agents with modular
low-dimensional subspace injection---three task families, six
allocation schemes, a credit-noise sweep, paired seeds, exact sign-flip
tests---and find no statistically detectable improvement over uniform
allocation in any on-pool comparison (no gain of $\geq 2$~pp). The
joint soft+$\sigma$ routing scheme is bounded within a $\pm 0.02$
equivalence margin in AUC
(time-averaged success) on 1.5B and 3B; concentrating the whole budget
on the credit argmax is marginally equivalent on 1.5B, where that
module is the verified bottleneck, and significantly worse on 3B;
inverse-propensity debiasing does not rescue routing. Misrouting has a
real cost (up to $-0.074$ AUC in-house; up to $-0.118$ end-to-end on
the BFCL-derived family). Across six fixed-step schedules the
loss is linear in the bottleneck's \emph{starvation rate} ($R^2 =
0.94$, descriptive), and a preregistered credit-free coverage floor
removes the detected harm. A matched-budget burst schedule and a
step-compensating catch-up arm are consistent with the harm residing in
insufficient \emph{cumulative parameter movement} of the bottleneck
rather than in update frequency. Our primary estimand is optimization
efficiency on a fixed task pool. A held-out evaluation on unseen BFCL
functions yields the study's one exception: soft routing exceeds
uniform on held-out endpoints ($+0.047$, $p = 0.031$, $n = 6$). A
plausible but untested reading is that the caller improvements favored
by routing are what transfer, while uniform's on-pool gains sat in a
synthesizer behavior specific to our harness (relaying a constant gold
answer); we report the exception rather than absorb it. We also document
three failure modes that silently invalidate ZO/ES experiments on
frozen LLMs.
\end{abstract}

\section{Introduction}

Zeroth-order (ZO) and evolutionary methods are increasingly used to
fine-tune frozen LLMs where backpropagation is unavailable or
memory-prohibitive \citep{malladi2023mezo, es_at_scale2025}. For
\emph{agents}---models that plan, select tools, call them, and synthesize
answers---recent work assigns blame for failed trajectories to specific
functional modules using verifiable signals: wrong tool selected, call
failed, final answer mismatched \citep{evotool2026, eglsca2026}. This
suggests an attractive composition: use trajectory-level credit to decide
\emph{where} the ZO/ES perturbation budget goes, concentrating search on
the module that is actually failing. We work in ZO/ES because the
agent's reward is an environment-verified, non-differentiable signal
(tool execution and exact-match answers) and because the question we
ask---how to split a fixed perturbation budget among modules---is
internal to ZO; we do not compare against gradient-based fine-tuning,
which would need a differentiable surrogate for that reward.

\begin{figure}[t]
\centering
\begin{subfigure}[b]{0.48\linewidth}
  \includegraphics[width=\linewidth]{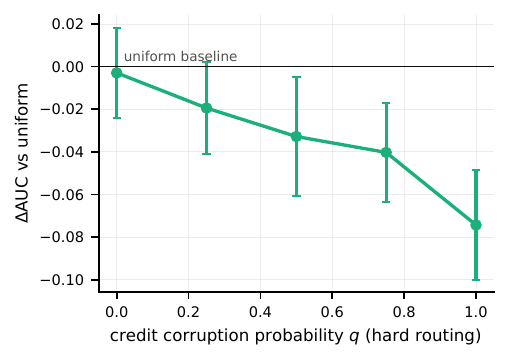}
  \caption{Credit-noise dose--response.}
  \label{fig:noise}
\end{subfigure}\hfill
\begin{subfigure}[b]{0.44\linewidth}
  \includegraphics[width=\linewidth]{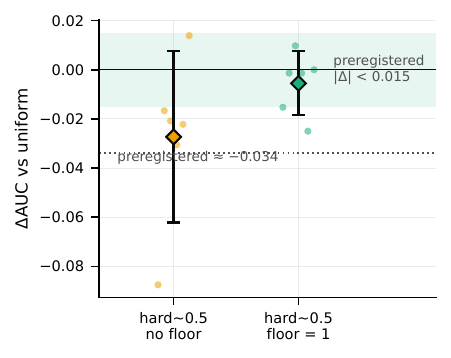}
  \caption{Preregistered floor test.}
  \label{fig:floor}
\end{subfigure}
\caption{(a) Corrupting the routing signal with probability $q$ degrades
hard routing monotonically; the endpoints $q = 0$ and $q = 1$ are the
hard and random arms of Table~\ref{tab:main}. (b) A credit-free coverage floor removes 80\%
of hard$\sim$0.5's penalty (a ratio with a noisy denominator;
\S\ref{sec:floor}), landing inside the shaded band, which marks
the \emph{preregistered prediction} $|\Delta| < 0.015$ (narrower than
the $\pm 0.02$ TOST margin).}
\end{figure}

This paper is a controlled test of that composition. Our answer is
negative, and the negative is informative:

\begin{enumerate}
\item \textbf{No detectable upside on the optimization pool.}
  The joint soft+$\sigma$ routing scheme shows no statistically
  detectable improvement over uniform in any headline on-pool
  comparison; a separate plain-soft, floored control also shows no
  upside (one held-out endpoint exception is reported in
  \S\ref{sec:heldout}). On
  Qwen2.5-1.5B a preregistered 12-seed replication bounds soft routing
  inside the $\pm 0.02$ equivalence margin by TOST ($\Delta = -0.010$; the
  pooled 20-seed estimate, $\Delta = -0.012$, 95\% CI $[-0.020, -0.003]$,
  is descriptive because the extension was decided after seeing the
  first eight seeds); on Qwen2.5-3B $\Delta = -0.003$ with TOST
  equivalence; at doubled budget $\Delta = -0.017$ with 5/5 seeds
  negative. A variant that concentrates 97\% of the budget on the module
  verifiably causing 7/8 residual failures is marginally equivalent to
  uniform on 1.5B ($p = 0.05$) and significantly worse on 3B.
\item \textbf{Real downside.} Concentrating on a random module costs
  $-0.074$ AUC ($p = 0.008$). Corrupting the routing signal with
  probability $q \in \{0.25, 0.5, 0.75\}$ degrades the point estimates
  monotonically in $q$ (Figure~\ref{fig:noise}).
\item \textbf{One variable organizes six schemes.} Across the six
  allocation schemes the model can score, the loss is linear in the
  \emph{bottleneck starvation rate} (the probability that the true
  bottleneck receives zero pairs in a generation): $\Delta\mathrm{AUC}
  \approx -0.001 - 0.087\cdot \mathrm{starvation}$, $R^2 = 0.94$
  (Figure~\ref{fig:starvation}); we present this as a descriptive
  regression whose inferential support is a within-seed fit and a
  seed-level bootstrap (\S\ref{sec:starvation}).
\item \textbf{Preregistered coverage test.} The starvation model predicts
  that a credit-free coverage floor ($\geq$1 pair per module per
  generation) eliminates the penalty. With predictions committed before
  the runs, the floored variant of a misrouted scheme (hard$\sim$0.5)
  recovered 80\% of a penalty that is itself not significant at $n = 6$
  (criterion: $\geq 60\%$) and is TOST-equivalent to uniform
  (Figure~\ref{fig:floor}).
\item \textbf{Cumulative movement is a candidate mediator.} A
  burst schedule that matches uniform's per-cycle budget but starves the
  bottleneck in time reproduces the harm ($-0.075$); a catch-up arm
  designed to compensate missing cumulative movement---while keeping
  the once-per-cycle update frequency---recovers uniform's level
  (\S\ref{sec:robust}). Under the self-normalized ES update, extra pairs
  buy variance reduction rather than larger steps, so in the schedules we
  tested, the result is consistent with insufficient cumulative movement
  mediating harm, and the starvation rate is the proxy that tracks it.
\item \textbf{One held-out exception.} Evaluating the trained
  coefficients on unseen BFCL functions, soft routing beat uniform on
  end-to-end success ($+0.047$, $p = 0.031$, $n = 6$) and the point
  estimates of all three routing arms' argument accuracy exceeded
  uniform's (only soft's difference is significant); we report this as
  the boundary of the on-pool finding (\S\ref{sec:heldout}).
\end{enumerate}

The prescriptive summary: \emph{in the failure-credit routing we
test---modular ZO/ES on frozen LLMs with tied low-dimensional
subspaces---maintaining coverage is the robust default; the tested
schedules are consistent with coverage helping by preserving each
module's cumulative movement. On the optimization pool, failure-credit
routing (unweighted or inverse-propensity-debiased) adds risk without
detectable reward}.
Per-layer untied
parameterizations and gain-estimating (bandit-style) credit remain
untested, and we scope the claim accordingly. This is consistent with
the small accuracy deltas (${\approx}1$~pp) that most adaptive-ZO work
reports over uniform baselines while its real gains are wall-clock and
memory---deltas of that size lie below what our $\pm 0.02$ margin can
certify, so ``no upside'' here means no upside of $\geq 2$~pp
(\S\ref{sec:related}, \S\ref{sec:setup}).

We also contribute a methodological record. Two intermediate conclusions
drawn during this study---a concentration dose--response relationship and
an analogy-based prediction for a sparse baseline---were later falsified
by our own controls, and are documented as such (\S\ref{sec:falsified},
\S\ref{sec:starvation}). The final claims survive a preregistered test
rather than post-hoc fitting.

\section{Related work}
\label{sec:related}

\paragraph{ZO/ES fine-tuning of LLMs.} MeZO \citep{malladi2023mezo}
established memory-efficient ZO fine-tuning; scaled ES variants optimize
full parameter sets \citep{es_at_scale2025, eggroll2025}. Low-rank and
subspace approaches include random subspaces \citep{subzero2024} and ES
over SVD singular values of adapters \citep{essa2025}, which our
parameterization follows. A unified subspace-perturbation analysis
\citep{subspace_theory2025} argues that a broad class of perturbation
structures shares convergence rates under its assumptions (random
projections with expected isotropy)---loosely consistent with our
equivalence finding, though it does not cover state-dependent credit
routing and establishes no allocation-level equivalence.

\paragraph{Adaptive allocation in ZO.} AdaLeZO \citep{adalezo2026} selects
layers with a bandit; Dominant-Layer ZO \citep{dominantlayer2026} tunes a
single dominant layer; other work learns per-block perturbation variances
\citep{zofinetuner2025}. For these methods the headline gains are
efficiency (1.7--4.5$\times$ wall-clock), and the reported accuracy
deltas over uniform/full baselines are small (${\approx}{+}0.1$ to
${\approx}{+}1.2$ pp across models)---an order-of-magnitude analogy only,
since endpoint accuracy and our time-averaged success are not directly
comparable, and since deltas of that size lie below the $\pm 0.02$
margin our equivalence certificates can exclude (\S\ref{sec:setup}).
Sparse MeZO \citep{sparsemezo2024} is different in kind: it reports
accuracy gains of up to ${\approx}9$ pp, but against full-perturbation
MeZO rather than a uniform allocation across units, and its mechanism is
parameter-level sparsity rather than routing a budget among modules; it
is therefore neither covered nor contradicted by our results. Note that
all of these methods select among tens to millions of units with a
budget far smaller than the unit count, a regime in which per-unit
coverage floors are infeasible; our design sits in the opposite regime
(Scope, \S\ref{sec:discussion}). AdaLeZO additionally applies inverse-propensity weighting to debias its
adaptive sampling; we test the same debiasing on our routing arm in
\S\ref{sec:robust} (it does not help), so our negative conclusion covers
inverse-propensity-debiased as well as unweighted credit routing. With
that scope, the results are
complementary rather than contradictory: in our regime, concentration
bought no sample efficiency, and the compute benefit those methods
report does not require a credit signal.

\paragraph{Credit assignment in LLM agents.} EvoTool \citep{evotool2026}
and EGL-SCA \citep{eglsca2026} route blame to textual/program updates.
CoPES \citep{copes2026} decomposes parameters into subspaces for
cooperative ES, but not credit-driven; \citet{beyondcb2026} apply ZO to
LoRA parameters of frozen LLM agents with uniform perturbation. To our
knowledge no prior work routes parameter-space search budget by
trajectory credit; we test this bridge and report that it does not help
on-pool optimization, with a candidate mechanism.

\paragraph{Classical antecedents.} The ES update we use---antithetic
(mirrored) sampling with common random numbers---follows
\citet{salimans2017es} and mirrored sampling for evolution strategies
\citep{brockhoff2010mirrored}; adaptive-subspace ES such as Guided ES
\citep{maheswaranathan2019guided} and ASEBO \citep{choromanski2019asebo}
shape the perturbation distribution rather than routing a budget among
modules. Our coverage finding has a textbook counterpart in randomized
block-coordinate descent, whose convergence guarantees require every
block to be sampled with probability bounded away from zero
\citep{nesterov2012coordinate, richtarik2014iteration}: a module with
zero sampling probability in a generation is exactly a violated
lower bound. Our contribution is to measure this condition's
consequences empirically for failure-credit routing in LLM agents (a
linear penalty in the starvation rate; a floor that removes it) and to
trace the penalty to interrupted cumulative movement.

\section{Setup}
\label{sec:setup}

\paragraph{Agent and environment.} A tool-using agent solves single-tool
tasks (20 arithmetic tasks executed by a whitelisted-AST calculator, 12
knowledge lookups; answers environment-verifiable). A rollout is four
controlled decoding stages under a frozen instruct model with greedy
decoding (``generation'' below always means an ES generation):
\textbf{Planner} (one-sentence plan), \textbf{Selector} (choose a tool by
name), \textbf{Caller} (emit JSON arguments, up to 2 attempts with error
feedback; arguments are executed for real), \textbf{Synthesizer} (produce
the final answer from the tool output). Reward $= \mathbf{1}[\text{success}]
+ 0.15\,\mathbf{1}[\text{tool correct}] + 0.15\,\mathbf{1}[\text{call ok}]
- 0.03\,(\text{attempts}-1)$.

\paragraph{Modular parameter subspaces.} Each module $m$ owns a 32-dim
coefficient vector $\theta_m$. For every target linear layer ($q/v/o$
projections of all blocks; 84 target projections on 1.5B, 108 on 3B,
72 on SmolLM2-1.7B) we take the
layer's top singular triples $(u_i, s_i, v_i)$ and assign them to modules
round-robin. During module $m$'s decoding stage only, a forward hook
adds $\Delta y = c \cdot \sum_i \theta_m[i]\, s_i (x^\top v_i)\, u_i$.
Bases derive from stable seeds; checkpoints store only the 128
coefficients, and resume is bit-exact (verified by property tests).

\paragraph{Credit and routing.} A rule-based assigner converts each failed
trajectory into a normalized blame vector using only verifiable signals:
wrong tool $\to$ selector; failed calls with the correct tool $\to$
caller; correct tool, successful call, wrong answer $\to$ synthesizer.
The planner receives no observable signal (a structural asymmetry we
report). An EMA tracker yields credit $c \in \Delta^3$. Routing schemes
map $c$ to per-module pair counts (largest-remainder rounding):
\textbf{uniform}; \textbf{soft} ($\mathrm{softmax}(\beta c)$, $\beta$
annealed $1\to3$, optionally with $\sigma_m$ scaled by credit,
``+$\sigma$''); \textbf{hard} (all free pairs to $\arg\max c$);
\textbf{random} (all pairs to one random module per generation);
\textbf{sparse} (equal split over a random half of the modules per
generation); and \textbf{hard$\sim q$} (hard, but with probability $q$
per generation the routing signal is replaced by a random vector). A
\emph{coverage floor} optionally guarantees each module $\geq 1$ pair.
All soft arms in the main comparisons use the credit-scaled-$\sigma$
variant (soft+$\sigma$ in tables, ``soft'' in prose); the debiasing
check of \S\ref{sec:robust} uses plain soft routing with a floor.

\paragraph{ES update.} Antithetic mirrored ES per module with common
random numbers across each pair
\citep{salimans2017es, brockhoff2010mirrored}; the per-generation update norm is
clipped at 0.5 per module (trust region; \S\ref{sec:traps}). Fitness $=$
task reward $+\; 0.5 \times$ mean teacher-forced log-probability of gold
targets (correct tool name / gold arguments / gold answer---all
environment ground truth, no LLM judge).

\paragraph{Protocol.} All arms share seeds and environment randomness
(common random numbers), a paired design enabling exact sign-flip
permutation tests ($2^n$ enumeration).\footnote{With $n = 6$ the smallest
attainable two-sided $p$ is $2/2^6 = 0.031$ (all six paired differences
sharing sign); such values denote maximal sign concordance at that sample
size, not a near-threshold effect. Key conclusions replicate across
backbones and task families rather than resting on any single test.}
Primary metric: time-averaged
success---the area under the success-vs-rollouts curve (not ROC-AUC);
per-generation rollout cost is identical across arms, so it equals mean
per-generation success. Equivalence claims use TOST with paired
Student-$t$ ($\mathrm{df} = n-1$)\footnote{Reported as the pair of
one-sided $p$-values $p_{\mathrm{upper}}/p_{\mathrm{lower}}$, where
$p_{\mathrm{upper}}$ tests that the mean falls below $+0.02$ and
$p_{\mathrm{lower}}$ that it exceeds $-0.02$; equivalence at level
$\alpha$ requires both below $\alpha$.} and a $\pm 0.02$ margin,
specified before the follow-up seed extension and all later experiments
(but after the first eight main-experiment seeds had been seen). The
margin is a practical-importance threshold, not a resolution limit:
$0.02$ AUC is ${\approx}0.3$ successes per generation at the 16-task
batch, less than one task's worth of per-generation change, and we
regard smaller differences as practically unimportant here. It is
\emph{larger} than the typical accuracy deltas (${\approx}1$~pp) reported
by adaptive-ZO methods over uniform baselines, so our certificates
exclude gains of $\geq 2$~pp and say nothing about smaller ones; we
state ``no upside'' in that sense throughout. Unless a caption states
otherwise, $\pm$ denotes SD across seeds. All randomness derives
from sha256-keyed streams; every results file records the full
configuration and git commit.

\paragraph{Estimand.} Each generation evaluates the center and every ES
candidate on a batch of 16 tasks drawn (seed-keyed) from the same
32-task pool, which is reused across generations, and the shaping term
uses the pool's gold targets. The primary metric therefore measures
\emph{optimization efficiency on a fixed task pool}---how quickly an
allocation scheme improves the agent on the tasks it is being optimized
on---not generalization to held-out examples. The cross-family
replications (\S\ref{sec:crossfamily}, \S\ref{sec:bfcl}) change the task
family, not the train/test relationship. To check that the learned
coefficients transfer at all, \S\ref{sec:heldout} evaluates the trained
coefficients of every BFCL run on 96 unseen BFCL functions.

\section{Three silent failure modes}
\label{sec:traps}

Each of the following produced a plausible-looking but invalid experiment
before being fixed; we report them as prerequisites for ZO/ES studies on
frozen LLMs.

\paragraph{T1: random subspace bases are behaviorally inert.} With
unit-norm random rank-1 bases on the last 4 layers, the logit gap at the
decisive output token was 7.5, closing at ${\sim}0.01$/generation;
scaling learning rate and injection strength $40\times$ still produced
\emph{zero} behavioral change in 15 generations. A full 24-run study ran
to completion with \textbf{bit-identical success curves in all arms}. SVD
singular-direction bases move the same decision point by $\pm 1.4$ logits
per $\sigma$-perturbation. This contrast confounds direction quality with
scale: random bases are unit-norm rank-1 while SVD directions carry their
singular values $s_i$, so part of the $40\times$-resistant inertness
reflects smaller effective magnitude, not near-orthogonality alone. We
therefore report T1 as an empirical failure to obtain \emph{any}
behavioral signal from unit-norm random bases at practical injection
scales, not as a magnitude-matched contrast; the actionable lesson---verify
that a basis actually moves behavior before trusting a null---is
unchanged.

A preregistered magnitude-matched follow-up settles the
confound at the working injection configuration (\S\ref{sec:robust},
ledger row~23): under the full 28-layer/32-dim setup used throughout,
learning is basis-\emph{direction}-agnostic---random-direction subspaces
match SVD ($\Delta\mathrm{AUC} = +0.008$, $p = 0.75$, $n = 6$) and
\emph{magnitude-matched} random directions equal or slightly exceed them
($+0.035$, $p = 0.06$; and beat unmatched random by $+0.027$,
$p = 0.031$, 6/6). Direction is not the active ingredient, magnitude contributes, and
the unit-norm inertness above is a scale effect. We accordingly claim only that
injection needs a per-module low-dimensional subspace at adequate
magnitude, and attach no direction-specific role to SVD; the
preregistered prediction that matching failed (T1 a pure direction
effect) is recorded as a miss.

\paragraph{T2: strong bases + unbounded steps diverge.} With SVD bases,
raw fitness differences explode when behavior collapses, a positive
feedback loop (one module's $\|\theta\|$ reached 21.8; success collapsed
to 0 without recovery). A per-module trust region removes this failure
mode. The region is two-sided: tightening it from $0.5$ to $0.125$ per
module costs $-0.072$ AUC by truncating the informative large-fitness
steps (\S\ref{sec:confounds}).

\paragraph{T3: greedy decoding makes task reward piecewise-constant
(basis-dependent).} Under weak random bases, mirrored pairs at
$\sigma = 0.3$ flipped any output token in 0 of 18 trials, so the two-point
gradient estimate is identically zero; a verifiable log-likelihood
shaping term restores a followable gradient without any LLM judge. With
strong SVD bases this term turns out to be \emph{optional}: a
preregistered necessity prediction failed (\S\ref{sec:shaping}), and
pure task reward suffices. We flag the trap because whether it binds
depends on basis quality, which is easy to get wrong (T1).

\section{Results}
\label{sec:results}

\paragraph{Roadmap.} \S\ref{sec:mock} is the synthetic pre-study.
\S\ref{sec:main}--\S\ref{sec:budget} establish the main negative result,
a falsified intermediate conclusion, and the cross-model, cross-family,
and budget replications (intro items 1--2).
\S\ref{sec:starvation}--\S\ref{sec:shaping} develop the starvation
regression, the preregistered floor test, and shaping robustness (items
3--4). \S\ref{sec:bfcl}--\S\ref{sec:lockm} move to the BFCL-derived
family, including the held-out exception (item 6) and the lock-$m$
dissociation. \S\ref{sec:confounds}--\S\ref{sec:robust} are
preregistered confound and robustness checks, including the
burst/catch-up analysis behind item 5; \S\ref{sec:synthesis}
synthesizes.

\subsection{Synthetic pre-study}
\label{sec:mock}

Before the LLM experiments we built a deterministic synthetic environment
with the same four-module pipeline, in which each module's success
probability is a smooth function of its own subspace and the true
bottleneck is known---permitting a genuine \emph{oracle} routing arm.
Across four regimes (a stationary bottleneck with a tight budget; a
non-stationary bottleneck with wide and with tight budgets; and the
non-stationary/wide regime repeated at $d = 32$; 10 seeds each,
paired): oracle routing never
outperformed uniform (all $|\Delta| \leq 0.009$ AUC, min $p = 0.11$);
soft routing showed no detectable advantage over uniform either (and was significantly
\emph{worse} in one regime, $\Delta = -0.006$, $p = 0.008$); while
wrongly-concentrated budgets cost up to $-0.220$
($p = 0.002$). These results anticipated the LLM findings below and fixed
the control arms of the main study. Full per-regime tables are in
Appendix~\ref{app:prestudy}.

\subsection{ES works; routing does not help}
\label{sec:main}

\begin{figure}[t]
\centering
\begin{subfigure}[b]{0.48\linewidth}
  \includegraphics[width=\linewidth]{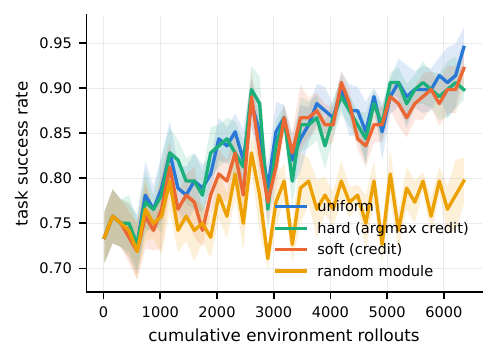}
  \caption{Learning curves (mean $\pm$ s.e., $n=8$).}
  \label{fig:curves}
\end{subfigure}\hfill
\begin{subfigure}[b]{0.48\linewidth}
  \includegraphics[width=\linewidth]{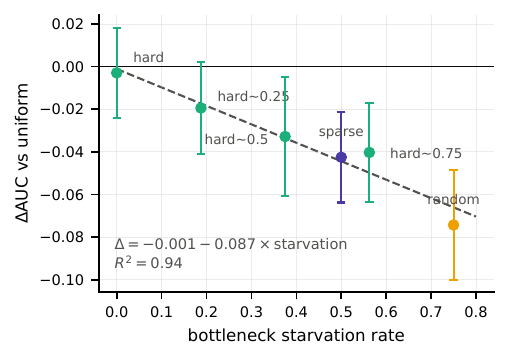}
  \caption{Starvation regression across six schemes.}
  \label{fig:starvation}
\end{subfigure}
\caption{(a) Arms optimize a frozen Qwen2.5-1.5B agent from success
0.734 to 0.90--0.92 (last-five-generation means; uniform reaches
${\approx}0.92$, final generation 0.945); only the misrouted arm lags. (b) Loss vs
uniform is linear in the bottleneck starvation rate ($R^2=0.94$); error
bars are 95\% CIs of paired differences.}
\end{figure}

Table~\ref{tab:main} and Figure~\ref{fig:curves} show the main comparison on Qwen2.5-1.5B (45
generations, 4 mirrored pairs/generation, batch 16, $n = 8$ seeds).

\begin{table}[htbp]
\centering
\caption{Main experiment (Qwen2.5-1.5B, $n=8$, paired; $\pm$ is SD
across seeds in this table). ``Conc.''\ is the mean per-generation
fraction of budget on the most-funded module. TOST column: \checkmark\
equivalence certified within $\pm 0.02$; \ding{55} not certified at
$n = 8$ (soft is certified on the 12-seed replication below); --- not
applicable. Hard's equivalence is marginal under paired-$t$ TOST (weaker
one-sided $p = 0.0497$).}
\label{tab:main}
\begin{tabular}{lccccc}
\toprule
arm & conc. & AUC & $\Delta$ vs uniform & $p$ (perm.) & TOST $\pm0.02$ \\
\midrule
uniform & 0.25 & $0.8418 \pm 0.0250$ & --- & --- & --- \\
soft+$\sigma$ & 0.64 & $0.8280 \pm 0.0145$ & $-0.0139$ & 0.156 & \ding{55} \\
hard ($\arg\max$ credit) & 1.00 & $0.8389 \pm 0.0215$ & $-0.0030$ & 0.766 & \checkmark \\
random module & 1.00 & $0.7675 \pm 0.0075$ & $\mathbf{-0.0743}$ & \textbf{0.008} & --- \\
\bottomrule
\end{tabular}
\end{table}

The first eight seeds left soft's sign ambiguous ($\Delta = -0.0139$,
$p = 0.156$). We preregistered a 12-seed extension (criteria fixed in
advance: pooled mean in $[-0.03, 0]$, upper CI $< +0.01$). Analyzed on
their own as a replication, the 12 new seeds are directionally consistent
but not individually significant ($\Delta = -0.0102$, 7/12 negative,
$p = 0.078$) while passing $\pm 0.02$ TOST ($p_{\mathrm{lower}} = 0.040$,
$p_{\mathrm{upper}} < 0.001$).
Because the extension was decided after seeing the first eight seeds, we
treat the pooled estimate ($n = 20$: $\Delta = -0.0117$, 95\% CI
$[-0.0204, -0.0029]$, pooled $p = 0.013$) as descriptive rather than a
fresh confirmatory test. The peek-independent conclusion is the
equivalence---soft's effect is bounded within $\pm 0.02$, TOST holding on
the 12-seed replication alone: no upside and at most a small bounded
downside.

The decisive contrast is \textbf{hard vs random}: identical concentration,
only the position differs; $\Delta = +0.0714$, $p = 0.008$, 8/8 seeds.
Credit attribution itself is \emph{correct}: hard allocates 97.2\% of
pairs to the synthesizer, which accounts for 7/8 residual failures at
$\theta = 0$. Being right buys nothing over uniform; being wrong is
expensive.

\subsection{A falsified intermediate conclusion}
\label{sec:falsified}

With only uniform/soft/random, AUC appears monotone in concentration
(per-seed regression slope $-0.099$, 8/8 negative, $p = 0.008$). Re-testing
on the position-correct triple (uniform/soft/hard) collapses the slope to
$-0.005$ ($p = 0.703$): the apparent dose--response in concentration was an
artifact of a single point. We report this to illustrate why a
control that decouples ``dose'' from co-varying factors (here: hard $=$
same dose, correct position) is mandatory before claiming a
dose--response.

\subsection{Cross-model replication and its boundary}
\label{sec:3b}

On Qwen2.5-3B ($n = 6$, 30 generations): soft remains equivalent to
uniform ($\Delta = -0.003$; paired-$t$ TOST $p = 0.002/0.006$); random remains significantly
worse ($\Delta = -0.055$, $p = 0.031$, 6/6 negative). One result does \emph{not}
transfer: hard is worse than uniform on 3B ($\Delta = -0.027$,
$p = 0.031$), with position still mattering directionally (hard beats
random in 5/6, $p = 0.16$). The single-bottleneck starvation model below does not
cover 3B, whose residual failures are spread over multiple modules and
whose credit concentrates on the caller rather than the synthesizer;
we call this a \emph{multi-module failure profile} (a description of
where failures and credit fall, not a measurement of where gain is
attainable) and report it as the model's boundary.

\subsection{Cross-family replication}
\label{sec:crossfamily}

To rule out a Qwen-specific artifact, we replicated the four main arms
on SmolLM2-1.7B-Instruct, a Llama-family architecture ($n = 6$, 30
generations). ES itself transfers: uniform lifts AUC to $0.544$ from a
zero-$\theta$ level of $0.406$. This model's credit profile is again
multi-module (selector-dominant), so the preregistered conditional
prediction applies: hard should underperform uniform. It does---$-0.095$
($p = 0.031$, 6/6 seeds)---a third failure configuration beyond the
synthesizer- and caller-dominant regimes above. Soft is within noise of
uniform ($+0.002$); random is directionally consistent but not
significant at this sample size ($-0.034$, negative in 4/6 seeds,
$p = 0.125$), a preregistered prediction that missed and is recorded as
such. No routing scheme outperformed uniform in a second model family.

\subsection{Budget scale}
\label{sec:budget}

Doubling the budget (4$\to$8 pairs, $n=5$) leaves soft's deficit intact
($-0.0172 \pm 0.0038$, 5/5 negative) and does not change uniform's own
performance ($\Delta = -0.005$, $p = 0.75$): budget is not the active
constraint in this regime, so this ablation cannot test the
exploration-insufficiency hypothesis (untested, not falsified).

\subsection{The starvation regression}
\label{sec:starvation}

Define the \emph{bottleneck starvation rate} as the probability that the
true bottleneck module receives zero pairs in a generation. Each scheme's
rate is analytic: $0$ for hard (the credit argmax sits on the
bottleneck), $3/4$ for random (the bottleneck is one of four
equally-likely-dropped modules), $1/2$ for sparse, and $q\cdot 3/4$ for
hard$\sim q$ (corruption reroutes to random with probability $q$); rates
measured from the allocation logs agree within $0.05$ (largest deviation
$0.043$ for hard$\sim$0.75; hard's measured rate is $0.028$ rather than
$0$ because largest-remainder ties occasionally move its argmax;
Appendix~\ref{app:starvation}). Across the six schemes the model can
score (hard, hard$\sim q$ at three levels, sparse, and random; uniform
is the reference and soft's mixed allocation has no single starvation
rate; Figure~\ref{fig:starvation}),
\[
\Delta\mathrm{AUC} \;=\; -0.001 \;-\; 0.087 \cdot \mathrm{starvation},
\qquad R^2 = 0.941 ,
\]
and refitting on the measured rather than nominal rates gives slope
$-0.091$, $R^2 = 0.973$. The intercept is zero: schemes that never
starve the bottleneck pay nothing. The six points share the same uniform
seeds and four of them come from one parameter family (hard$\sim q$), so
the pooled $R^2$ is descriptive; the inferential evidence is that
fitting the six schemes \emph{within each seed} gives slope $-0.087 \pm
0.032$ (8/8 negative, sign-flip $p = 0.008$), that a seed-level
bootstrap (10{,}000 resamples) puts the pooled slope's 95\% CI at
$[-0.108, -0.066]$, and that leave-one-scheme-out prediction errors are
$\leq 0.005$ for four of six schemes (worst: random, $0.017$). Competing
variables fail: expected pairs-on-bottleneck ($R^2 = 0.82$) cannot
separate random from sparse (both $E = 1$, observed $0.032$ apart);
corruption probability cannot score sparse at all. We present the
\emph{coverage} reading (zero-starvation schemes pay nothing; starvation
is what costs) as the robust claim and the linear coefficient as
exploratory; \S\ref{sec:robust} shows that the starvation rate is itself
a proxy for the bottleneck's cumulative parameter movement.

\emph{Preregistered miss (recorded).} Before the sparse arm ran we
predicted $\Delta \in [-0.035, -0.015]$ by analogy to hard$\sim$0.5; the
observation ($-0.0425$) fell outside. The miss exposed the correct
variable: sparse's starvation is 0.50 (not 0.375), and the starvation
model retrodicts it at $-0.044$.

\subsection{Preregistered test of the coverage floor}
\label{sec:floor}

If starvation is the mechanism, a credit-free floor ($\geq 1$ pair per
module) must eliminate the penalty. Predictions were committed to the
repository before the runs (6 pairs/generation, $n = 6$): un-floored
hard$\sim$0.5 $\approx -0.034$; floored hard$\sim$0.5 $|\Delta| < 0.015$
(i.e., the floor removes $\geq 60\%$ of the penalty).

\begin{table}[htbp]
\centering
\caption{Floor test vs preregistered predictions ($\pm$ is SD across
seeds).}
\label{tab:floor}
\begin{tabular}{lcc}
\toprule
arm & predicted & observed \\
\midrule
hard$\sim$0.5, no floor & $\approx -0.034$ & $-0.0273 \pm 0.0332$ (5/6 neg.) \\
hard$\sim$0.5, floor $=1$ & $|\Delta| < 0.015$ & $\mathbf{-0.0056 \pm 0.0124}$, TOST $p{=}0.002/0.018$ \\
\bottomrule
\end{tabular}
\end{table}

The floor removes \textbf{80\%} of the penalty. The 80\% ratio has a
noisy denominator, however: the in-house un-floored penalty
($-0.0273$, $p = 0.062$, 5/6 negative) is itself not significant at
$n = 6$. We therefore rest the mechanistic reading on the floored arm's
TOST-certified equivalence to uniform and on the BFCL replication
(Figure~\ref{fig:bfcl}), where both the un-floored failure ($-0.107$,
$p = 0.031$) and the floor's difference-in-differences rescue ($+0.133$,
95\% CI $[+0.121, +0.145]$, $p = 0.031$) are significant. The floor is
thus a demonstrated robustness lever; because it co-varies several
quantities at once (\S\ref{sec:confounds}), we present coverage as the
sufficient intervention we could identify, not as the sole isolated
cause.

\subsection{Shaping-term robustness}
\label{sec:shaping}

At doubled shaping weight ($w = 1.0$, $n=6$): random remains
significantly worse ($\Delta = -0.073$, $p = 0.031$) and hard's point
estimate is $+0.001$ (no detectable difference; equivalence not
certifiable at this $n$ due to higher variance). At $w = 0$ (uniform
only, $n = 3$; preregistered prediction: shaping necessary, AUC
$< 0.78$), the
prediction \emph{failed}: uniform reaches $0.841 \pm 0.016$ with clearly
rising learning curves, versus $0.862$ for $w = 0.5$ on the same three
seeds (paired $\Delta = -0.021$, $n = 3$; the $n = 8$ mean is $0.842$).
Uniform therefore optimizes without the shaping term---the
preregistered necessity prediction is falsified---though the same-seed
comparison leaves room for a contribution of the ${\approx}2$~pp order
that $n = 3$ cannot resolve; the routing comparisons
replicate at $w = 1.0$ ($w = 0$ was run for uniform only, as a necessity
test). The shaping term is not necessary for learning given SVD
bases---defusing the concern that the uniform baseline's optimization
depended on it.

\subsection{External validity: a standard function-calling benchmark}
\label{sec:bfcl}

We build a \emph{BFCL-derived} agent evaluation: 48 tasks sampled
(fixed seed, function-name deduplicated) from BFCL~v3 \emph{simple}
\citep{bfcl2024}, with argument verification following the official
checker semantics. This is not the official BFCL protocol: our harness
adds per-task distractor functions for the selector, caller retries with
error feedback, and a synthesizer stage. One consequence must be stated
plainly: BFCL tasks have no natural-language answer, so the verified
tool output instructs the model to relay the constant string
\texttt{CORRECT}, which is every task's gold answer. The synthesizer
stage therefore learns a task-independent relay, whereas the caller must
emit task-specific structured arguments. We treat this harness as a
\emph{diagnostic environment} with a standard task distribution, not as
the official benchmark, and we report the caller-only accuracy (the
quantity the official AST checker scores) in the held-out evaluation of
\S\ref{sec:heldout}. Under an exploratory 96-token
harness, zero-$\theta$ success is $0.521$ with residual failures spread
across caller, synthesizer, and selector---a \emph{multi-module} headroom
profile, like the 3B setting and unlike the single-bottleneck in-house
tasks; under the experiment configuration (32-token decoding), the
in-sweep zero-$\theta$ level (generation-0 evaluations) is $0.500$.
Preregistered predictions: misrouting penalty transfers; \emph{hard}
loses to uniform (the multi-module boundary phenomenon, \S\ref{sec:3b});
soft shows no upside.

\begin{figure}[t]
\centering
\includegraphics[width=0.62\linewidth]{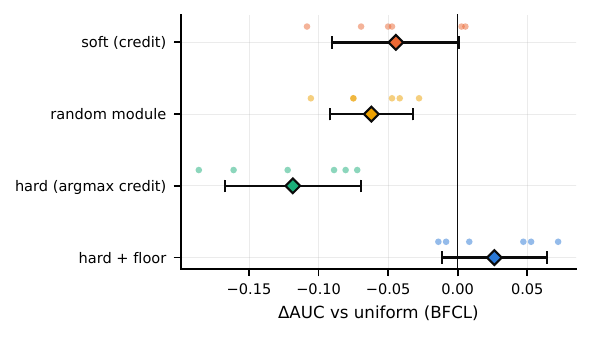}
\caption{BFCL-derived task family ($n=6$, paired; diamonds: mean
with 95\% CI; dots: per-seed differences; soft/random/hard at
4 pairs/generation, hard$+$floor at 6; its DiD baseline, the un-floored
6-pair hard arm at $-0.107$, is not drawn). Every routing scheme
underperforms uniform---hard most severely ($-0.118$, $p=0.031$, net
learning $+0.013$ vs uniform's $+0.132$)---and a credit-free coverage
floor removes the detected harm ($+0.026$ vs uniform, n.s.). End-to-end
success on the training pool; the arm gaps are carried by the
synthesizer relay (\S\ref{sec:heldout}).}
\label{fig:bfcl}
\end{figure}

Results (Figure~\ref{fig:bfcl}; end-to-end success on the training
pool---\S\ref{sec:heldout} shows that these arm differences are carried
by the synthesizer relay rather than by argument accuracy): random
$-0.062$ ($p = 0.031$, 6/6)
\emph{hit}; hard $\mathbf{-0.118}$ ($p = 0.031$, 6/6; net progress over
the in-configuration zero-$\theta$ level a mere $+0.013$, versus
uniform's $+0.132$) \emph{hit}; soft $-0.044$ fell below its predicted band
$[-0.03, +0.01]$ (a miss we record: no upside confirmed, magnitude worse
than predicted). Unpredicted and noteworthy: hard $<$ random
($-0.057$, $p = 0.031$)---locking onto the credit argmax (the caller)
is worse than rotating at random because hard starves the synthesizer
every generation while random does not (\S\ref{sec:heldout} shows the
synthesizer relay carries this gap). At the time we read this as a hint
that a multi-module generalization of the starvation regression should
weight starvation by \emph{achievable gain} rather than failure counts
(tested in \S\ref{sec:lockm}).

The floor test transfers (preregistered, pairs $= 6$): un-floored hard
replicates the failure ($-0.107$, $p = 0.031$, 6/6; predicted
$\leq -0.06$); adding the floor shifts hard by a paired
difference-in-differences of $+0.133$ (95\% CI $[+0.121, +0.145]$,
$p = 0.031$), landing at $+0.026$ vs uniform ($p = 0.219$)---no
detectable difference, with a positive point estimate that $n = 6$
cannot resolve (equivalence not certified here; certified in the
single-bottleneck regime, Table~\ref{tab:floor}). The floor thus removes
the detected harm in \emph{both} headroom regimes.

\subsection{Held-out evaluation of the trained coefficients}
\label{sec:heldout}

All results above are measured on the task pool being optimized. To
test whether the learned coefficients transfer, we took every BFCL run's
saved coefficients and evaluated them---no further training---on 96
BFCL~v3 \emph{simple} tasks whose questions \emph{and} function names
are disjoint from the 48 training tasks (distractors drawn within the
held-out set; same harness). The held-out functions share BFCL's API
families and naming conventions with the training set, so this tests
near-distribution transfer rather than out-of-family generalization.
Because checkpoints retain only each run's
final three generations, this is an endpoint evaluation (generation~30
primary; the mean over generations 28--30 as a robustness read), not a
held-out learning curve. We report end-to-end success and the
caller-only accuracy (the argument check the official AST checker
scores, unaffected by the synthesizer's constant relay). Predictions
were committed before the replay (\texttt{a928d05}): uniform beats the
untrained coefficients on held-out tasks by $\geq 0.05$ (R1); the arm
ordering replicates, hard $\leq -0.03$ vs uniform (R2); and arm
differences are carried by caller accuracy while the synthesizer relay
rate $P(\text{success} \mid \text{call ok})$ differs by $\leq 0.05$
across arms (R3).

\begin{table}[htbp]
\centering
\caption{Held-out BFCL evaluation (96 unseen functions; generation-30
coefficients; $n = 6$, $\pm$ SD across seeds). Relay is
$P(\text{success} \mid \text{call ok})$, the synthesizer's rate of
relaying the constant answer. Bold marks the best value per column
(point estimates; only soft's differences from uniform are individually
significant). The untrained row has no seed variance because
$\theta = 0$ is deterministic under greedy decoding.}
\label{tab:heldout}
\begin{tabular}{lccc}
\toprule
arm & success & caller accuracy & relay \\
\midrule
untrained ($\theta = 0$) & $0.583$ & $0.698$ & $0.836$ \\
uniform & $0.696 \pm 0.031$ & $0.696 \pm 0.031$ & $1.000 \pm 0.000$ \\
soft+$\sigma$ & $\mathbf{0.743} \pm 0.016$ & $\mathbf{0.745} \pm 0.014$ & $0.998 \pm 0.006$ \\
random & $0.717 \pm 0.022$ & $0.727 \pm 0.019$ & $0.986 \pm 0.022$ \\
hard & $0.660 \pm 0.043$ & $0.733 \pm 0.025$ & $0.901 \pm 0.059$ \\
\bottomrule
\end{tabular}
\end{table}

\paragraph{Results (Table~\ref{tab:heldout}).} Transfer holds (R1,
hit): uniform's trained coefficients raise held-out success from the
untrained $0.583$ to $0.696$ ($+0.113$, 6/6 seeds). But the transfer is
carried entirely by the synthesizer: uniform's held-out caller accuracy
is unchanged ($0.696$ vs $0.698$ untrained) while its relay rate rises
from $0.836$ to $1.000$---the task-independent constant relay of
\S\ref{sec:bfcl} transfers when learned. The on-pool arm
ordering does \emph{not} replicate (R2, miss). Hard remains the worst
arm in end-to-end success ($-0.036$ vs uniform, 5/6 negative,
$p = 0.25$; that sub-criterion held), but the random-below-uniform
sub-criterion failed ($+0.021$, 4/6 positive, $p = 0.34$), and soft
routing---not addressed by R2---\emph{exceeds} uniform on held-out
success ($+0.047$, 6/6 positive, $p = 0.031$; $+0.040$, $p = 0.062$ on
the 28--30 mean). No single module carries all arm gaps: soft's
advantage is caller-carried, whereas hard's end-to-end deficit is
synthesizer-carried (R3, miss, in the direction the harness caveat
anticipated). The point estimates of all three routing arms'
held-out caller accuracy exceed uniform's (soft $+0.049$, $p = 0.031$;
hard $+0.036$, $p = 0.062$; random $+0.031$, $p = 0.062$; only soft
significant), while hard's relay rate is $0.099$ below uniform's (5/6
negative)---hard starved the synthesizer, which then fails to relay,
and that alone accounts for its end-to-end deficit. That the credit-free
random arm also shows a (non-significant) caller gain weakens any
credit-specific reading, but the three routing arms do not share a
budget-share mechanism: soft ($0.51$) and hard ($0.96$) spend more of
their budget on the caller than uniform's fixed quarter, while random's
realized caller share is $0.24$---by construction it matches uniform's
quarter in expectation, and what distinguishes it is temporal
concentration (a few full-budget, lower-variance caller updates rather
than one pair every generation), not share.

Three readings follow, and we adopt all three. First, on this harness
uniform's on-pool advantage over hard (\S\ref{sec:bfcl}) is a
synthesizer-relay effect: the module uniform keeps covered learns a
constant that generalizes trivially, and starving it costs end-to-end
success on any task. This supports the concern that the constant relay
shapes the BFCL arm ordering, so the BFCL end-to-end numbers should not
be read as evidence about function-calling ability; by the held-out
evidence, this harness has a multi-module failure profile with a single
trivially attainable gain (the relay), so ``regime'' labels in this
paper describe where failures and credit fall, not where gain is
attainable.

Second, on the quantity the official benchmark
scores---argument accuracy on unseen functions---the routing arms'
point estimates are slightly \emph{better} than uniform's: soft and
hard spend more budget on the caller (random matches uniform's share
but concentrates it into fewer, fuller updates), caller improvements
are the ones that transfer, and uniform's
held-out caller accuracy is statistically flat at the untrained level
(we did
not log per-arm on-pool caller accuracy, so we cannot say how much
uniform's caller improved on the pool).

Third, and most important for
the headline: this is the
one comparison in the study in which a credit-routed scheme beats
uniform. It is on a held-out endpoint metric rather than on the
optimization-efficiency estimand, unpredicted (the preregistered
ordering prediction missed), on one task family, at $n = 6$, at the
$p = 0.031$ resolution floor. We report it as an exception to the ``no
upside'' finding rather than absorbing it, and discuss what it suggests
in \S\ref{sec:discussion}.

\subsection{Attainable gain dissociates from credit}
\label{sec:lockm}

Why is following credit so costly on BFCL? Lock-$m$ probes that
concentrate the entire budget on one fixed module ($n = 4$, 30
generations) separate the two quantities a credit router conflates.
Credit mass concentrates on the caller ($0.80$), yet the only module
whose locked optimization climbs by a nontrivial margin is the
synthesizer: $+0.129$ AUC over the in-configuration zero-$\theta$ level
($0.500$), and paired $+0.146 \pm 0.020$ (4/4 seeds) over the
lock-planner/lock-selector trajectories, which are behaviorally inert
and so track $\theta = 0$ on the same batches; exploratory at this $n$.
Lock-planner and lock-selector produce bit-identical trajectories in
4/4 seeds: those two subspaces are behaviorally inert on this task
family. Credit tracks where failures
\emph{occur}; progress requires perturbing where gain is
\emph{attainable}; on this harness's on-pool metric these are different
modules, so even a perfectly accurate failure signal misallocates the
budget for that metric. Two caveats
bound this reading. First, it is exploratory at $n = 4$. Second, the
harness shapes it: the synthesizer's target is the constant relay
described in \S\ref{sec:bfcl}, plausibly the easiest gain available,
while the caller's task-specific arguments are the hardest---so the
dissociation may be partly manufactured by the harness rather than a
property of function-calling agents in general. The held-out evaluation
(\S\ref{sec:heldout}) is consistent with the second caveat: the synthesizer's
``attainable gain'' is the constant relay, which transfers trivially,
whereas the caller gains that credit routing buys are the ones that
transfer to unseen functions. We therefore keep the dissociation out of
the headline claims and use it only to motivate the gain-weighting test
below. Our
preregistered composite test of a gain-weighted starvation regression
nonetheless missed---per-module gain estimates are noisy, and near-zero
estimates corrupt the weights---which we record. An exploratory
synthesizer-only weighting reproduces the observed arm ordering
(uniform $>$ soft $>$ random $>$ hard) in 4/4 pairwise comparisons; we
label it post hoc.

\subsection{Preregistered checks of two design confounds}
\label{sec:confounds}

Two features of the setup could in principle manufacture arm differences
without any starvation. We preregistered a test for each.

\paragraph{Basis binding.} Modules are bound to singular-direction
groups round-robin, so module identity correlates with basis strength.
We reran the four main arms with the module-to-group binding permuted by
a seed-keyed shuffle ($n = 6$, 45 generations; four predictions
preregistered). All four hit: the arm ordering replicates exactly
(uniform $>$ hard $>$ soft $>$ random); random $-0.079$ ($p = 0.031$,
6/6; predicted $\leq -0.04$ and significant); hard $-0.005$ (predicted
$|\Delta| < 0.02$; equivalence not certifiable at $n = 6$); soft
$-0.028$ (predicted $\leq +0.01$). Binding strength cannot explain any
between-arm difference.

\paragraph{Trust-region capacity.} Arms that update fewer modules per
generation have a smaller total step budget ($4 \times 0.5$ for uniform
versus $1 \times 0.5$ for hard). We preregistered that a uniform arm
with per-module clip $0.125$ (total $0.5$, matched to hard) would stay
within $|\Delta| < 0.02$ of uniform, reasoning from the budget-scale
null that capacity is not the active constraint. The prediction
\emph{missed}: capacity-matched uniform loses $-0.072$ ($p = 0.031$,
6/6; 95\% CI $[-0.107, -0.037]$), the deficit grows over training
(first half $-0.041$, second half $-0.105$), and it lands at random's
level ($+0.005$ vs random, n.s.)---while hard, at the same total clip,
beats it by $+0.069$ ($p = 0.031$, 6/6).

A post-hoc analysis (labeled as such) of realized step norms, recovered
from consecutive checkpoints over the final three generations of every
run (the checkpoint difference is the update actually applied, after
every scaling and clip), shows why the miss does not revive the
confound. At clip $0.5$ the
trust region does not bind late in training (0 of 274 module updates
across the four in-house sweeps; 5 of 205 on BFCL), with realized
per-module steps of $0.10$--$0.15$. Realized total movement per
generation is $0.37$ for uniform, $0.12$ for hard and $0.08$ for random:
hard moves a third as much as uniform and matches its AUC, while random
moves as much as hard and loses $0.07$. Total capacity does not track
outcome; \emph{which} module moves does. At clip $0.125$, by contrast,
the region binds in 33--58\% of late updates, truncating precisely the
large fitness-difference steps that carry signal while leaving
noise-sized steps intact---an over-tight region degrades the fitness
weighting ES relies on. The capacity experiment therefore identifies a
second edge of the T2 trap (the region must be neither absent nor
tight), not a capacity explanation of the arm ordering. An
instrumented replay of seed 1 (bit-identical trajectories, first 20
generations) extends the picture to early training: at clip $0.5$ the
region binds only for uniform (25\% of its synthesizer updates, whose
single-pair estimates carry the largest norm) and never for hard or
random (0 of 20 touched updates each), with realized per-generation
movement of $0.54$ for uniform, $0.14$ for hard and $0.10$ for
random---the clip constrains only the arm it is alleged to favor. At
clip $0.125$ it binds in 41\% of all module updates (75\% on the
synthesizer) from the first generation, halving realized movement to
$0.28$.

\subsection{Three preregistered robustness checks}
\label{sec:robust}

Three challenges to the claims---does the ``do not route'' conclusion
survive inverse-propensity debiasing; is T1 a direction effect or a scale
artifact; does the floor uniquely identify starvation as the mechanism---were
turned into preregistered experiments (predictions in
\texttt{97f92db}, before the runs; ledger rows 21--25). The catch-up
arm (Q3) was specified before it ran (\texttt{fbbd097}) as a
\emph{two-sided} diagnostic---either outcome was informative---so it is
reported as a confirmatory follow-up rather than scored as a prediction
(ledger row~26). All use the single-bottleneck in-house setting
(synthesizer bottleneck), 45 generations, $n = 6$; O and Q were
preregistered at $n = 8$ and run at $n = 6$ for compute reasons, a
deviation we disclose.

\paragraph{Debiasing does not rescue routing (O).} On a floored
soft-routing arm we added inverse-propensity weighting in the spirit of
AdaLeZO ($w_m = (1/M)/\pi_m$, with $\pi_m$ the realized allocation share
of module $m$ in that generation): the estimand is the
uniform-allocation update, and the weight is a no-op under uniform.
AdaLeZO's sampling-with-replacement estimator differs in detail, so
this is an isomorphic debiasing rather than a re-implementation; this
arm uses 8 pairs per generation so that a floor of 1 leaves 4 routable
pairs. The
floored soft arm is
already TOST-equivalent to uniform ($\Delta = -0.001$, equivalence
certified), and adding IPW does \emph{not} recover an upside: it loses
$-0.024$ versus uniform ($p = 0.031$, 6/6) and $-0.022$ versus the same
arm without weighting ($p = 0.031$). The mechanism is structural: each
module's mirrored-pair estimate is self-normalized (scale
$1/(\text{pairs}\cdot\sigma)$), so it carries no propensity bias for IPW
to remove; the weights merely amplify under-sampled modules' steps and
inject variance. The preregistered ``no gain'' criterion held (no arm
exceeded uniform by $>0.02$); the significant \emph{loss} was not
predicted and is reported as an unpredicted finding. This widens the negative
conclusion from unweighted to inverse-propensity-debiased routing.

\paragraph{Basis learning is direction-agnostic; magnitude is the active
variable (P).} To separate T1's direction-quality claim from injection
scale, we reran uniform routing with three per-module bases: SVD
singular directions (the default), unit-norm random directions, and
\emph{magnitude-matched} random directions (random unit vectors rescaled
to the corresponding SVD singular values). Mean AUC over 45 generations:
SVD $0.842$ (the $n = 8$ main-experiment uniform arm, reused as the
reference; $0.846$ on seeds 1--6), random $0.854$, magnitude-matched
random $0.881$ (all climbing from the untrained ${\approx}0.74$). Paired
over seeds 1--6: random matches SVD ($+0.008$, $p = 0.75$),
magnitude-matched equals or slightly exceeds SVD ($+0.035$, $p = 0.06$)
and significantly beats unmatched random ($+0.027$, $p = 0.031$, 6/6).
The preregistered prediction---that matching would fail and T1 was a
pure direction (alignment) effect---\emph{missed}: direction is not the
active ingredient (unit-norm random directions already match SVD), and
magnitude contributes (matched beats unmatched), so SVD directions are
not necessary. This is
consistent with, and resolves, the scale confound flagged for the T1
inertness result (\S\ref{sec:traps}); note that eval batches of 16
quantize success to $1/16$, so these are 6--8-seed AUC aggregates rather
than single-generation reads.

\paragraph{Starvation harm is consistent with insufficient cumulative
movement rather than update frequency (Q).} The coverage floor changes
several quantities at once, so
we isolated temporal starvation with a \emph{burst} schedule: every $P$th
generation gives all $P$ pairs to the synthesizer, other generations
rotate them among the remaining modules, matching per-cycle total pairs
and per-generation capacity to uniform while denying the bottleneck
temporal coverage (its starvation rate rises $0\to0.75$). Burst loses
$-0.075$ versus uniform ($p = 0.031$, 6/6), reproducing the starvation
harm at matched budget (Q1). Raising the burst arm's trust-region cap to
$P\times$ leaves the run \emph{bit-identical}: the clip never binds on
the synthesizer's burst-generation updates, the only updates the cap
change affects (0 of 72), so the harm is not a capacity-ceiling
artifact (Q2). Finally, because a self-normalized ES
buys variance-reduction rather than larger steps from extra pairs, burst
delivers the synthesizer far less \emph{cumulative} movement than
uniform; a \emph{catch-up} arm that scales each burst-generation step by
$P$ (designed to compensate cumulative movement while keeping the once-per-$P$ update
frequency) recovers the loss: $+0.076$ versus plain burst ($p = 0.031$,
6/6) and $+0.0005$ versus uniform (its point estimate sits at uniform's
level, though the $n = 6$
spread leaves strict $\pm 0.02$ equivalence uncertified) (Q3). This is
consistent with the starvation harm being mediated by cumulative
parameter displacement rather than by update recency: coverage matters
because uniform allocation is what lets every module accumulate enough
movement. We stop short of calling it an isolation, because the
catch-up arm also enlarges the individual burst-generation step
($4\times$) and we have no control showing that step enlargement alone
is neutral (\S\ref{sec:confounds} shows outcomes are sensitive to the
step regime), and because we do not distinguish signal-driven from
noise-driven displacement; a schedule that restores cumulative movement
through many small updates would close the first gap. Note also that
the catch-up arm lies off the starvation regression by construction
(starvation $0.75$, $\Delta \approx 0$ where the fit predicts
$-0.066$): the regression is scoped to schedules with uncompensated
steps.

\subsection{Synthesis}
\label{sec:synthesis}

Across experiments spanning 3 backbones in 2 model families, 3 task
families (synthetic, in-house tool-use, BFCL), 6 allocation schemes, 2
budget levels, a noise dose--response, and 2 shaping weights for the
routing comparisons (a third for uniform only),
\emph{uniform allocation was never significantly outperformed in any
on-pool comparison}. The one comparison in which a routing scheme beat
uniform is a held-out endpoint on BFCL (\S\ref{sec:heldout}). Five
preregistered robustness checks (basis binding, trust-region capacity,
inverse-propensity debiasing, basis magnitude, and a matched-budget
burst schedule) leave the arm ordering intact and are consistent with
the mechanism, two of their predictions having missed in ways that
refine rather than reverse it. The effect structure on the pool:
routing has no detectable upside (equivalence certified for soft
routing on 1.5B and 3B and for oracle-quality credit in three of four
synthetic regimes; marginal for hard concentration on 1.5B). Its downside is
$-0.087 \times$ starvation in the single-bottleneck regime and up to
$-0.118$ end-to-end with full concentration in the multi-module
failure regime (a gap \S\ref{sec:heldout} shows is carried by the
synthesizer relay on that harness). The remedy---a coverage
floor---requires no credit
signal, removes the detected harm in both regimes (preregistered, hit
twice; the BFCL hit is on the relay-mediated end-to-end metric), and
certifies equivalence in the single-bottleneck regime.

\section{Discussion and limitations}
\label{sec:discussion}

\paragraph{Why no upside?} With antithetic pairs and common random
numbers, a single mirrored pair already yields a usable directional
estimate in a 32-dim subspace (noisier than a multi-pair estimate, but
the self-normalized update does not let extra pairs enlarge the step);
per-module marginal returns therefore diminish too slowly for
reallocation to matter (consistent with the budget-scale null). The
binding resource is each module's cumulative displacement, which uniform
coverage supplies by default and which step compensation can also
supply (\S\ref{sec:robust}); basis magnitude matters for the same reason
(\S\ref{sec:robust}, P).

\paragraph{The one exception, and what it suggests.} On held-out BFCL
functions, soft routing beat uniform ($+0.047$, $p = 0.031$; $p = 0.062$
on the three-generation mean) and the point estimates of all three
routing arms' argument accuracy exceeded uniform's
(\S\ref{sec:heldout}). The pattern is coherent for the credit-routed
arms: soft and hard spend
more budget on the caller than uniform's fixed quarter; caller
improvements are what transfer to unseen functions; uniform spent a
quarter of its budget on a synthesizer whose learned behavior is a
task-independent constant, and its held-out caller accuracy is
statistically flat. The credit-free random arm matches uniform's caller
share (realized $0.24$) yet shows a similar non-significant caller
gain---so if the effect is real it need not be credit-specific, and
budget share cannot be the whole mechanism; what random adds over
uniform is concentrating its caller budget into fewer, fuller
updates. This suggests a hypothesis we did
not preregister and cannot test with the present data: credit routing
may help \emph{generalization} when the credit-targeted module is the
one whose improvements transfer, even while it buys nothing in
on-pool optimization efficiency. Testing it needs held-out learning
curves (our checkpoints retain only endpoints), a harness in which the
synthesizer's target is not a constant, and replication beyond one task
family and $n = 6$; the on-pool ``no upside'' finding and the
cumulative-movement mechanism stand, but the practical prescription
should be read as ``keep coverage; do not expect routing to speed
optimization,'' not as ``routing cannot help what the agent learns.''

\paragraph{Scope.}
Three backbones in two model families, three task
families, 30--45 generation horizons, greedy decoding. The quantitative
starvation regression is established on the single-bottleneck in-house
setting; on multi-module settings (3B, BFCL) its qualitative structure
holds (concentration harms, floor rescues) but the preregistered
gain-weighted generalization missed and only an exploratory
synthesizer-only weighting fits (\S\ref{sec:lockm}). The planner has no
observable credit signal. $n = 6$--$8$ seeds bounds the certifiable
equivalence margin at $\pm 0.02$, which is above the ${\approx}1$~pp
gains typical of the adaptive-ZO literature (\S\ref{sec:setup}). The
primary estimand is optimization efficiency on a fixed task pool; the
held-out check of \S\ref{sec:heldout} covers endpoints on one task
family only, and held-out learning curves were not measured.
Most headline soft arms jointly route pairs and credit-scaled $\sigma$;
plain soft routing is isolated only in the floored eight-pair IPW sweep.
Accordingly, cross-backbone equivalence identifies the joint
perturbation-budget scheme, not pair allocation alone.

\emph{Regime.} Every experiment has $M = 4$ modules and 4--8 pairs per
generation, so a per-module floor is cheap---at 4 pairs it \emph{is}
uniform, which is why the floor tests use 6. Adaptive-ZO methods route
among tens to millions of units with budgets far below the unit count;
there a per-unit floor is infeasible, starvation of most units every
step is unavoidable, and routing decides \emph{whom} to starve. Our
coverage prescription is not directly executable in that regime; the
candidate cumulative-movement reading (\S\ref{sec:robust}) suggests
that temporal rotation with step
compensation could substitute for per-step coverage, but that is an
untested extrapolation.

\emph{Multiple comparisons.} The paper reports roughly seventy
significance tests, many at the $p = 0.031$ resolution floor of $n = 6$;
we apply no family-wise correction, and under a global null about
$70 \times 0.031 \approx 2$ floor-level results would be expected by
chance. We distinguish preregistered directional tests (ledger,
Appendix~\ref{app:prereg}) from unpredicted findings (BFCL hard $<$
random, the IPW loss, the capacity miss, and the held-out soft
advantage), and we regard the latter's floor-level significance as
awaiting replication.
The ledger's hit count is descriptive bookkeeping over criteria of
uneven strictness, not a success rate.

The design-confound and robustness checks (\S\ref{sec:confounds},
\S\ref{sec:robust}, \S\ref{sec:heldout}) are summarized in
\S\ref{sec:synthesis}; two of their preregistered predictions missed
(capacity matching; basis direction), and the trust-region size is a
sensitive hyperparameter in its own right. Of twenty-eight scored preregistered predictions, nineteen hit and nine
missed; all are reported, alongside one inert manipulation (the
capacity-matched burst clip) recorded as a control and one two-sided
diagnostic (the catch-up arm) that is not scored. Finally, our claim is scoped to
\emph{failure}-credit routing---unweighted or
inverse-propensity-debiased---over tied low-dimensional subspaces.
Gain-estimating credit (bandit-style, as in AdaLeZO) and per-layer
untied parameterizations remain untested and not contradicted here; the
lock-$m$ dissociation names gain-based signals as a natural next test,
though the held-out replay warns that on-pool gain estimates can favor
task-independent behavior (the constant relay) over what transfers.

\paragraph{Relation to adaptive-ZO gains.} Nothing here contradicts
wall-clock or memory benefits of perturbing fewer parameters; our claim
concerns sample efficiency at fixed rollout budget. The two views compose
into one recommendation: choose subsets for systems reasons, keep
coverage guarantees where the budget permits them, and do not expect
credit signals to buy convergence.

\section{Conclusion}

On the optimization pool, trajectory-level credit can find the module
where an LLM agent's failures \emph{occur}, but spending ZO/ES
perturbation budget there bought nothing in optimization efficiency in
any setting we tested and carried a quantifiable downside whenever it
interrupted another module's cumulative movement. Uniform coverage---or, in the
routing families we tested, any scheme with a per-module floor---is the
robust default for optimizing on a task pool; whether routing toward
the module whose gains transfer can buy generalization, as one
held-out result suggests, is an open question we state rather than
settle. We offer the starvation
regression, its cumulative-movement reading, and three experimental
prerequisites as reusable guidance, and our
falsified intermediate conclusions as a case study in why decoupling
controls and preregistration matter for negative results.

\subsubsection*{Reproducibility statement}
All randomness derives from sha256-keyed streams; interrupted runs resume
bit-identically (property-tested). The anonymous supplement contains the
code, task files, preregistration record, and a compact per-seed,
per-generation results dataset covering the LLM experiments and the
synthetic pre-study, sufficient to recompute every statistic and figure
in this paper (an overview of every sweep is in
Appendix~\ref{app:overview}). Compact training-result files store each
run's full configuration, originating git commit, configuration hash,
and backend. The held-out replay file stores its preregistration commit,
source sweep, task identifiers, per-task outcomes, and the source runs'
configuration hashes; its analysis script reads this compact file
directly. The full run checkpoints (each run's final three generations,
learned coefficients, credit state, and complete per-generation history)
are not part of the anonymous archive and will be released with the
de-anonymized artifact.
Logged step norms are the updates actually applied after all scaling
and clipping (for the IPW and catch-up sweeps, whose logged norms
predate this instrumentation, the reported movements are recovered from
checkpoint differences). Software: Python 3.10.12, PyTorch 2.11.0+cu130,
Transformers 5.5.4; models: Qwen2.5-1.5B-Instruct (rev.\ 989aa79),
Qwen2.5-3B-Instruct (rev.\ aa8e725), SmolLM2-1.7B-Instruct
(rev.\ 31b70e2); greedy decoding throughout.
Hardware: one RTX 5080 Laptop GPU (16\,GB, $\leq 7$\,GB used); total
compute ${\approx}280$ GPU-hours including pilots, failed runs, and the
robustness checks of \S\ref{sec:robust}. Seeds: 1--8 (main, credit-noise, sparse), 9--20 (seed extension), 1--6
(other ablations, except $w = 0$: 1--3 and the doubled-budget arm:
1--5), 1--4 (lock-$m$ probes), 1--10 (synthetic pre-study).

\bibliography{main}
\bibliographystyle{tmlr}

\appendix

\section{Preregistration ledger}
\label{app:prereg}

Every prediction below was written to
\texttt{doc/experiments\_stage2.md} in the commit shown, which precedes
the commit that reported its outcome. The anonymous supplement includes
a hash/date/subject manifest of this commit order; the public repository
history will be released after review (dates are 2026). Nineteen of
twenty-eight scored predictions
hit; one manipulation (row~25) was inert and is recorded as a control,
and one follow-up (row~26) was specified as a two-sided diagnostic
before it ran and is not scored. Criteria vary in strictness (row~2 is
a point prediction without a tolerance band, and its observation is
itself not significant), so the count is bookkeeping, not a success
rate.

\begin{table}[htbp]
\centering
\small
\begin{tabular}{r>{\raggedright\arraybackslash}p{6.2cm}lll}
\toprule
\# & Preregistered prediction & Commit & Date & Outcome \\
\midrule
1  & Sparse: $\Delta \in [-0.035,-0.015]$ vs uniform$^\dagger$ & \texttt{3b87179} & 08-19 & miss ($-0.043$) \\
2  & Floor F1: un-floored hard$\sim$0.5 $\approx -0.034$ & \texttt{3b87179} & 08-19 & hit \\
3  & Floor F2: floored $|\Delta|<0.015$ (removes $\geq$60\%) & \texttt{3b87179} & 08-19 & hit \\
4  & Shaping G1: random ($w{=}1$) $\leq -0.04$ \& sig. & \texttt{11762e7} & 08-20 & hit \\
4b & Shaping G1: hard ($w{=}1$) $|\Delta|<0.02$ & \texttt{11762e7} & 08-20 & hit ($+0.001$; TOST n.c.) \\
5  & Shaping G2: shaping necessary ($w{=}0$ AUC $<0.78$) & \texttt{11762e7} & 08-20 & miss (0.841) \\
6  & BFCL H: random $\leq -0.03$ \& sig. & \texttt{7e65bd4} & 08-20 & hit \\
7  & BFCL H: hard $<0$ (multi-module) & \texttt{7e65bd4} & 08-20 & hit \\
8  & BFCL H: soft $\in [-0.03,+0.01]$ & \texttt{7e65bd4} & 08-20 & miss ($-0.044$) \\
9  & BFCL floor I1: un-floored hard $\leq -0.06$ & \texttt{5248e38} & 08-21 & hit \\
10 & BFCL floor I2: floor removes $\geq$50\% & \texttt{5248e38} & 08-21 & hit \\
11 & Power J: $n{=}20$ soft mean $\in[-0.03,0]$, upper CI $<+0.01$ & \texttt{3e35a93} & 08-22 & hit \\
12 & Cross-family K: hard $<$ uniform (multi-module, cond.) & \texttt{8408094} & 08-22 & hit \\
13 & Cross-family K: random significantly worse & \texttt{3e35a93} & 08-22 & miss ($p{=}0.125$) \\
14 & Cross-family K: soft $\leq +0.01$ & \texttt{3e35a93} & 08-22 & hit \\
15 & Gain-weighted L: gain-weighted ordering 4/4 & \texttt{3e35a93} & 08-22 & miss \\
16 & Binding M: arm ordering replicates (permuted) & \texttt{90ad693} & 08-22 & hit \\
17 & Binding M: random $\leq -0.04$ \& sig. & \texttt{90ad693} & 08-22 & hit \\
18 & Binding M: hard $|\Delta|<0.02$ & \texttt{90ad693} & 08-22 & hit \\
19 & Binding M: soft $\leq +0.01$ & \texttt{90ad693} & 08-22 & hit \\
20 & Capacity N: matched uniform $|\Delta|<0.02$ & \texttt{90ad693} & 08-22 & miss ($-0.072$) \\
21 & IPW O1: floored soft $|\Delta|<0.02$ vs uniform & \texttt{97f92db} & 08-25 & hit \\
22 & IPW O2: debiased routing no upside ($\leq +0.01$) & \texttt{97f92db} & 08-25 & hit ($-0.024$)$^\ddagger$ \\
23 & Basis P1: matched random $\ll$ SVD (direction) & \texttt{97f92db} & 08-25 & miss ($+0.035$) \\
24 & Burst Q1: burst $\leq -0.02$ vs uniform, majority negative & \texttt{97f92db} & 08-25 & hit ($-0.075$, 6/6) \\
25 & Burst Q2: cap-matched burst $|\Delta|$ vs burst & \texttt{97f92db} & 08-25 & inert$^\S$ \\
26 & Catch-up Q3: two-sided diagnostic (recover $\Rightarrow$ movement; not $\Rightarrow$ frequency)$^\P$ & \texttt{fbbd097} & 08-27 & recovered ($+0.076$), not scored \\
27 & Held-out R1: uniform $-$ untrained $\geq +0.05$ & \texttt{a928d05} & 08-28 & hit ($+0.113$; via relay) \\
28 & Held-out R2: hard $\leq -0.03$ \& $\geq$4/6 neg.; random majority neg. & \texttt{a928d05} & 08-28 & miss (random $+0.021$, 4/6 pos.) \\
29 & Held-out R3: caller carries arm gaps; relay $|\Delta|\leq0.05$ & \texttt{a928d05} & 08-28 & miss (gaps via relay $-0.099$, not caller) \\
\bottomrule
\end{tabular}
\caption{Preregistration ledger (28 scored: 19 hit, 9 miss; row~25 inert,
reported as a control; row~26 unscored). $^\ddagger$O2's ``no upside''
criterion held; the significant \emph{loss} was unpredicted and is
reported as such (\S\ref{sec:discussion}, multiple comparisons). $^\S$The capacity-matched burst clip never binds, so
the arm is bit-identical to plain burst---an inert manipulation, not a
scored hit/miss. $^\P$The catch-up arm was specified before it ran as a
diagnostic whose two outcomes had opposite readings, without a
directional prediction; O and Q were preregistered at $n = 8$ and run at
$n = 6$. $^\dagger$The sparse
prediction was an informal analogy recorded in the analysis commit rather
than a separately time-stamped preregistration; it is counted as a miss.}
\label{tab:prereg}
\end{table}

\section{Starvation regression: raw points}
\label{app:starvation}

The six regression points of \S\ref{sec:starvation}
(Figure~\ref{fig:starvation}), each an in-house sweep at $n = 8$ paired
against the same uniform seeds. The fit
$\Delta\mathrm{AUC} = -0.001 - 0.087\cdot\mathrm{starvation}$
($R^2 = 0.941$) uses the analytic (nominal) starvation rate; the
empirically measured rate from the allocation logs is shown alongside for
audit.

\begin{table}[htbp]
\centering
\small
\begin{tabular}{lccccc}
\toprule
scheme & starv.\ (nominal) & starv.\ (measured) & $\Delta\mathrm{AUC}$ & 95\% CI & $p$ \\
\midrule
hard          & 0.000 & 0.028 & $-0.003$ & $[-0.024,+0.018]$ & 0.766 \\
hard$\sim$0.25 & 0.188 & 0.200 & $-0.019$ & $[-0.041,+0.002]$ & 0.070 \\
hard$\sim$0.5  & 0.375 & 0.353 & $-0.033$ & $[-0.061,-0.005]$ & 0.023 \\
sparse        & 0.500 & 0.500 & $-0.043$ & $[-0.064,-0.021]$ & 0.016 \\
hard$\sim$0.75 & 0.562 & 0.519 & $-0.040$ & $[-0.063,-0.017]$ & 0.008 \\
random        & 0.750 & 0.767 & $-0.074$ & $[-0.100,-0.049]$ & 0.008 \\
\bottomrule
\end{tabular}
\caption{Starvation regression points ($n = 8$ each, paired sign-flip
$p$). Nominal and measured starvation rates agree within $0.05$
(deviations $0.028$, $0.012$, $0.022$, $0.000$, $0.043$, $0.017$);
refitting on measured rates gives slope $-0.091$, $R^2 = 0.973$.}
\label{tab:starvation}
\end{table}

\section{Synthetic pre-study: the four regimes}
\label{app:prestudy}

Deltas vs uniform on the deterministic synthetic environment
(\S\ref{sec:mock}), 10 seeds and 120 generations per regime, paired.
Oracle routing (true bottleneck known) shows no detectable advantage over
uniform in any regime
($|\Delta| \leq 0.009$); wrongly-concentrated random routing costs up to
$-0.220$.

\begin{table}[htbp]
\centering
\small
\begin{tabular}{lccccc}
\toprule
regime & uniform AUC & oracle & soft & hard & random \\
\midrule
stationary, tight        & 0.900 & $-0.009$ & $-0.004$ & $-0.037^*$ & $-0.141^*$ \\
non-stationary, wide     & 0.821 & $-0.001$ & $-0.006^*$ & $-0.003$ & $+0.003$ \\
non-stationary, tight    & 0.768 & $-0.006$ & $-0.007$ & $-0.032^*$ & $-0.220^*$ \\
non-stationary, wide ($d{=}32$) & 0.767 & $+0.001$ & $+0.001$ & $-0.010$ & $+0.001$ \\
\bottomrule
\end{tabular}
\caption{Synthetic pre-study, $\Delta\mathrm{AUC}$ vs uniform. $^*$paired
sign-flip $p < 0.05$. Oracle vs uniform passes $\pm 0.02$ TOST in three
of four regimes ($p_{\mathrm{upper}}/p_{\mathrm{lower}}$: ${<}0.001/0.021$,
${<}0.001/{<}0.001$, $0.006/0.057$, ${<}0.001/{<}0.001$, in row order;
the non-stationary/tight regime is not certified). Per-seed values are
in the released results dataset.}
\label{tab:prestudy}
\end{table}

\section{Experiment overview}
\label{app:overview}

Table~\ref{tab:overview} lists every sweep behind a reported number:
backbone, task pool, arms, seeds ($n$), ES generations, mirrored pairs
per generation, coverage floor, shaping weight $w$, per-module clip, and
where it is reported. Unless stated otherwise: Qwen2.5-1.5B-Instruct,
32-dim modules on SVD bases, $\sigma = 0.3$, learning rate $0.3$,
evaluation batch 16 (12 on BFCL), greedy decoding. Every sweep is a
paired design over the listed seeds.

\begin{table}[htbp]
\centering
\footnotesize
\setlength{\tabcolsep}{2.8pt}
\begin{tabular}{ll>{\raggedright\arraybackslash}p{4.1cm}cccccl}
\toprule
sweep & backbone / tasks & arms & $n$ & gens & pairs & floor & $w$ & reported \\
\midrule
synthetic pre-study & mock, 4 regimes & uniform, oracle, soft, hard, random & 10 & 120 & 4 / 8 & 0 & --- & \S\ref{sec:mock}, App.~\ref{app:prestudy} \\
main & 1.5B / in-house & uniform, soft+$\sigma$, hard, random & 8 & 45 & 4 & 0 & 0.5 & \S\ref{sec:main} \\
seed extension & 1.5B / in-house & uniform, soft+$\sigma$ & 12 & 45 & 4 & 0 & 0.5 & \S\ref{sec:main} \\
credit noise & 1.5B / in-house & hard$\sim$0.25, 0.5, 0.75 & 8 & 45 & 4 & 0 & 0.5 & \S\ref{sec:starvation} \\
sparse & 1.5B / in-house & uniform, sparse & 8 & 45 & 4 & 0 & 0.5 & \S\ref{sec:starvation} \\
floor & 1.5B / in-house & uniform, hard$\sim$0.5, hard$\sim$0.5+floor & 6 & 45 & 6 & 0 / 1 & 0.5 & \S\ref{sec:floor} \\
budget scale & 1.5B / in-house & uniform, soft+$\sigma$ & 5 & 45 & 8 & 0 & 0.5 & \S\ref{sec:budget} \\
shaping $w{=}1$ & 1.5B / in-house & uniform, hard, random & 6 & 45 & 4 & 0 & 1.0 & \S\ref{sec:shaping} \\
shaping $w{=}0$ & 1.5B / in-house & uniform & 3 & 45 & 4 & 0 & 0 & \S\ref{sec:shaping} \\
cross-model & 3B / in-house & uniform, soft+$\sigma$, hard, random & 6 & 30 & 4 & 0 & 0.5 & \S\ref{sec:3b} \\
cross-family & SmolLM2-1.7B / in-house & uniform, soft+$\sigma$, hard, random & 6 & 30 & 4 & 0 & 0.5 & \S\ref{sec:crossfamily} \\
BFCL & 1.5B / BFCL-48 & uniform, soft+$\sigma$, hard, random & 6 & 30 & 4 & 0 & 0.5 & \S\ref{sec:bfcl} \\
BFCL floor & 1.5B / BFCL-48 & uniform, hard, hard+floor & 6 & 30 & 6 & 0 / 1 & 0.5 & \S\ref{sec:bfcl} \\
BFCL held-out & 1.5B / BFCL-96 unseen & replay of the BFCL arms' final coefficients & 6 & --- & --- & --- & --- & \S\ref{sec:heldout} \\
lock-$m$ & 1.5B / BFCL-48 & lock-planner, -selector, -caller, -synthesizer & 4 & 30 & 4 & 0 & 0.5 & \S\ref{sec:lockm} \\
binding permuted & 1.5B / in-house & uniform, soft+$\sigma$, hard, random & 6 & 45 & 4 & 0 & 0.5 & \S\ref{sec:confounds} \\
capacity (clip 0.125) & 1.5B / in-house & uniform & 6 & 45 & 4 & 0 & 0.5 & \S\ref{sec:confounds} \\
step-norm replay & 1.5B / in-house & uniform, hard, random; uniform at clip 0.125 & 1 & 20 & 4 & 0 & 0.5 & \S\ref{sec:confounds} \\
IPW (O) & 1.5B / in-house & uniform, soft, soft+IPW & 6 & 45 & 8 & 1 & 0.5 & \S\ref{sec:robust} \\
basis (P) & 1.5B / in-house & uniform on random / magnitude-matched bases & 6 & 45 & 4 & 0 & 0.5 & \S\ref{sec:robust} \\
burst (Q) & 1.5B / in-house & uniform, burst-synth., +cap-match, +catch-up & 6 & 45 & 4 & 0 & 0.5 & \S\ref{sec:robust} \\
\bottomrule
\end{tabular}
\caption{Experiment overview. ``BFCL-48'' is the 48-task training pool;
``BFCL-96 unseen'' the held-out set with disjoint questions and function
names. Per-module clip is $0.5$ except where noted. Synthetic regimes
use 4 pairs (tight) or 8 (wide) and module dimensions 16, 8, 16, 32 in
the row order of Table~\ref{tab:prestudy}.}
\label{tab:overview}
\end{table}

\end{document}